\pdfoutput=1

\documentclass[11pt]{article}

\usepackage{EACL2023}

\usepackage{times}
\usepackage{latexsym}

\usepackage[T1]{fontenc}

\usepackage[utf8]{inputenc}

\usepackage{microtype}

\usepackage{inconsolata}

\usepackage{verbatim}
\usepackage{graphicx}
\usepackage{caption}
\usepackage{subcaption}
\title{Automatically Summarizing Evidence from Clinical Trials: \\ A Prototype Highlighting Current Challenges}

\author{
  Sanjana Ramprasad\\
  Northeastern University\\
  \texttt{ramprasad.sa@northeastern.edu}
  \\\And
  Denis Jered McInerney\\
  Northeastern University\\
  \texttt{mcinerney.de@northeastern.edu}
  \\\AND
  Iain J. Marshall\\
  King's College London\\
  \texttt{iainjmarshall@gmail.com}
  \\\And
  Byron C. Wallace\\
  Northeastern University\\
  \texttt{b.wallace@northeastern.edu}
}

\begin{document}
\maketitle
\begin{abstract}

We present \emph{TrialsSummarizer}, a system that aims to automatically summarize evidence presented in the set of randomized controlled trials most relevant to a given query.
Building on prior work \cite{marshall2020trialstreamer}, the system retrieves trial publications matching a query specifying a combination of condition, intervention(s), and outcome(s), and ranks these according to sample size and estimated study quality.
The top-$k$ such studies are passed through a neural multi-document summarization system, yielding a synopsis of these trials.
We consider two architectures: A standard sequence-to-sequence model based on BART \cite{lewis2019bart}, and a multi-headed architecture intended to provide greater transparency to end-users.
Both models produce fluent and relevant summaries of evidence retrieved for  queries, but their tendency to introduce unsupported statements render them inappropriate for use in this domain at present.
The proposed architecture 
may help users verify outputs allowing users 
to trace generated tokens back to inputs. 
The demonstration video is available at: \url{https://vimeo.com/735605060}
The prototype, source code, and model weights are available at: \url{https://sanjanaramprasad.github.io/trials-summarizer/}. 
\end{abstract}

\section{Introduction}
\label{section:intro}
\begin{figure*}
    \centering
    \frame{\includegraphics[scale=0.27]{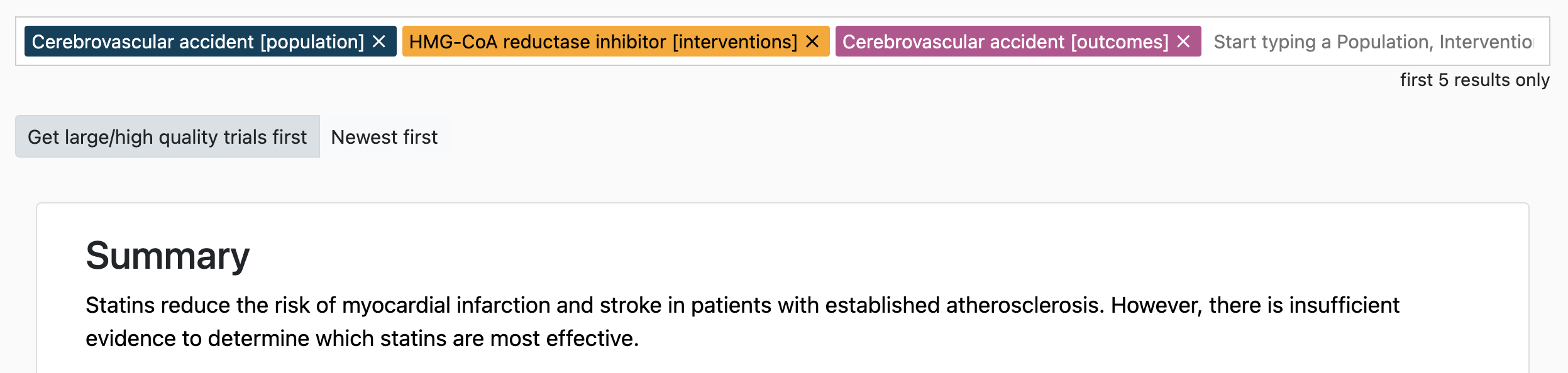}}
    \caption{An example query (regarding use of \emph{statins} to reduce risk of \emph{stroke}) and output summary provided by the system. In this example, the summary accurately reflects the evidence, but this is not always the case.}
    \label{fig:fig-1}
    \vspace{-.5em}
\end{figure*}

Patient treatment decisions would ideally be informed by all available relevant evidence.  
However, realizing this aim of evidence-based care has become increasingly difficult as the medical literature (already vast) has continued to rapidly expand \citep{bastian-10}.
Well over 100 new RCT reports are now published every day \citep{marshall2021state}.
Language technologies --- specifically automatic summarization methods --- have the potential to provide concise overviews of all evidence relevant to a given clinical question, providing a kind of \emph{systematic review} on demand \cite{wang-etal-2022-overview,deyoung2021ms2,wallace2021generating}.






We describe a demonstration system, \emph{TrialsSummarizer}, which combines retrieval over clinical trials literature with a summarization model to provide narrative overviews of current published evidence relevant to clinical questions. 
Figure \ref{fig:fig-1} shows an illustrative query run in our system and the resultant output.
A system capable of producing \emph{accurate} summaries of the medical evidence on any given topic could dramatically improve the ability of caregivers to consult the whole of the evidence base to inform care. 

However, current neural summarization systems are prone to inserting inaccuracies into outputs \citep{kryscinski2019evaluating,maynez2020faithfulness,pagnoni2021understanding,ladhak2021faithful,choubey2021mofe}.
This has been shown specifically to be a problem in the context of medical literature summarization  \cite{wallace2021generating,otmakhova-etal-2022-patient}, 
where there is a heightened need for factual accuracy. 
A system that produces plausible but often misleading summaries of comparative treatment efficacy is useless without an efficient means for users to assess the validity of outputs.



Motivated by this need for transparency 
when summarizing clinical trials, we implement a summarization architecture and interface designed to permit interactions that might instill trust in outputs.
Specifically, the model associates each token in a generated summary with a particular source ``aspect'' extracted from inputs.
This in turn allows one to trace output text back to (snippets of) inputs, permitting a form of verification. 
The architecture also provides functionality to ``in-fill'' pre-defined \emph{template summaries}, providing a compromise between the control afforded by templates and the flexibility of abstractive summarization.
We realize this functionality in our system demonstration.

\vspace{-.4em}
\section{Related Work}
\vspace{-.4em}

The (lack of) factuality of neural summarization systems is an active area of research \cite{chen2021improving, cao2020factual, dong2020multi, liu2020adapting, goyal2021annotating, zhang2021fine, kryscinski2019evaluating, xie2021factual}. 
This demo paper considers this issue in the context of a specific domain and application.
We also explored controllability to permit interaction, in part via templates. This follows prior work on hybrid template/neural summarization \cite{hua2020pair,mishra2020template,wiseman2018learning}.


We also note that this work draws upon prior work on visualizing summarization system outputs \cite{vig2021summvis,strobelt2018s,tenney2020language} and biomedical literature summarization \cite{plaza2013evaluating,demner2006answer,molla2010corpus,sarker2017automated, wallace2021generating}.
However, to our knowledge this is the first working prototype to attempt to generate (draft) evidence reviews that are both interpretable and editable on demand.

\begin{figure*}
    \centering
    \includegraphics[scale=0.39]{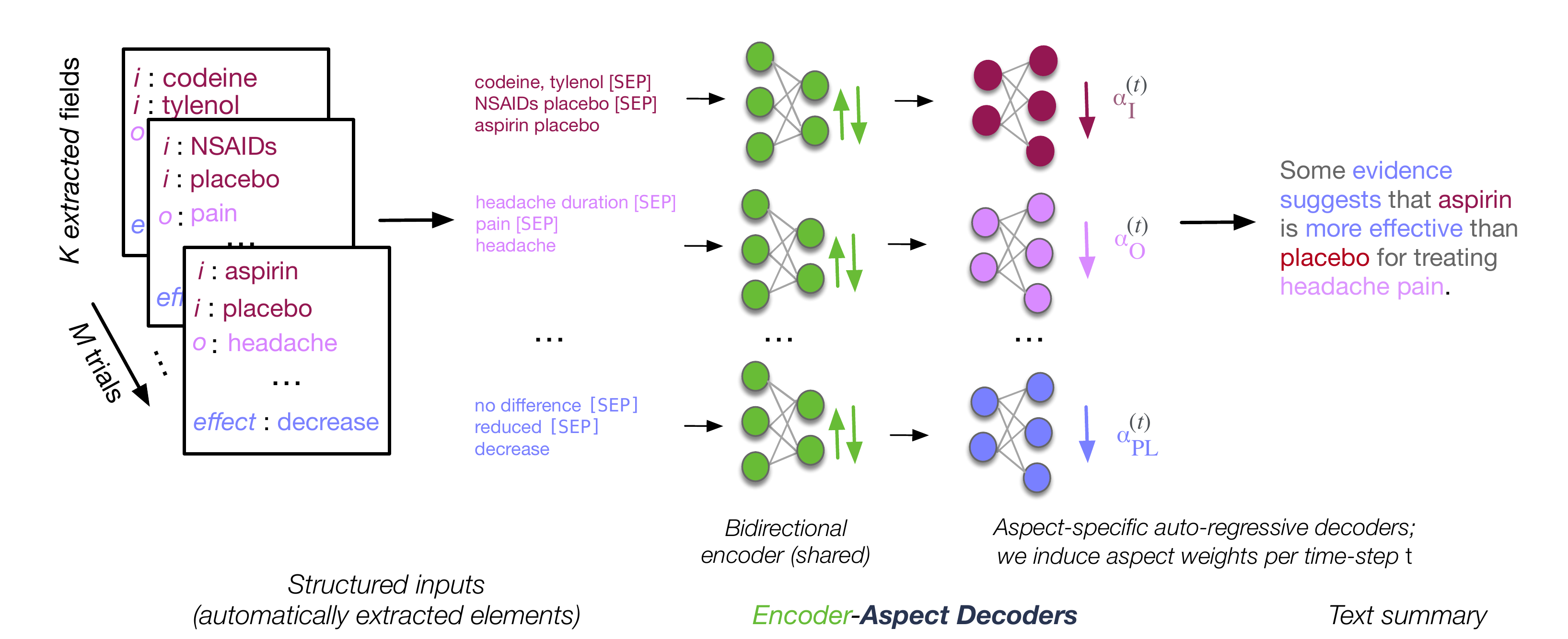}
    \caption{Our proposed structured summarization approach entails synthesizing individual aspects (automatically extracted in a pre-processing step), and conditionally generating text about each of these.}
    \label{fig:struct-sum}
    \vspace{-.5em}
\end{figure*}

\section{System Overview}
\label{section:system-design}

Our interface is built on top of Trialstreamer \cite{marshall2020trialstreamer}, an automated system that identifies new reports of randomized controlled trials (RCTs) in humans and then extracts and stores salient information from these in a database of all published trial information.  
Our system works by identifying RCT reports relevant to a given query using a straightforward retrieval technique (Section \ref{section:appendix_art_ret}), and then passing the top-$k$ of these through a multi-document summarization model (Section \ref{section:summarize}). 
For the latter component we consider both a standard sequence-to-sequence approach and a \textit{aspect structured} architecture (Section \ref{sec:structured_model}) intended to provide greater transparency.


\subsection{Retrieving Articles}
\label{section:appendix_art_ret}

Trialstreamer \cite{marshall2020trialstreamer,nye2020trialstreamer} monitors research databases --- specifically,  PubMed\footnote{\url{https://pubmed.ncbi.nlm.nih.gov/}} and the World Health Organization International Clinical Trials Registry Platform --- to automatically identify newly published reports of RCTs in humans using a previously validated classifier \cite{marshall2018machine}. 

Articles describing RCTs are then passed through a suite of machine learning models which extract key elements from trial reports, including: sample sizes; descriptions of trial populations, interventions, and outcomes; key results; and the reliability of the evidence reported (via an approximate risk of bias score; \citealt{higgins2019assessing}).
This extracted (semi-)structured information is stored in the Trialstreamer relational database.

Extracted free-text snippets describing study populations, interventions, and outcomes (PICO elements) are also mapped onto MeSH terms,\footnote{MeSH --- short for Medical Subject Headings --- is a controlled vocabulary maintained by the National Library of Medicine (NLM).} using a re-implementation of MetaMap Lite \cite{demner2017metamap}. 

To facilitate search, users can enter MeSH terms for a subset of populations, interventions, and outcomes, which is used to search for matches over the articles and their corresponding extracted key data in the database. 
Matched studies are then ranked as a score function of sample size $s$ and risk of bias score $\mathrm{rob}$: $\mathrm{score} = s / \mathrm{rob}$; that is, we prioritize retrieval of large, high-quality trial reports.


The novelty on offer in this system demonstration is the inclusion of a \emph{summarization} component, which consumes the top-$k$ retrieved trials (we use $k$=5 here) and outputs a narrative summary of this evidence in the style of a systematic review abstract \cite{wallace2021generating}.
By combining this summarization module with the Trialstreamer database, we can provide real-time summarization of all trials that match a given query (Figure \ref{fig:fig-1}).

\subsection{Summarizing Trials}
\label{section:summarize}

We consider two realizations of the summarization module. 
We train both models on a dataset introduced in prior work which comprises collections of RCT reports (PICO elements extracted from abstracts) as inputs and Authors' Conclusions sections of systematic review abstracts authored by members of the Cochrane Collaboration as targets \cite{wallace2021generating} (see Section \ref{sec:appendix_data_training}).

As a first model, we adopt BART \cite{lewis2019bart} with a Longformer \cite{beltagy2020longformer} encoder to accommodate the somewhat lengthy multi-document inputs.
As inputs to the model we concatenate spans extracted from individual trials containing salient information, including populations, interventions, outcomes, and ``punchlines.'' 
The latter refers to extracted snippets which seem to provide the main results or findings, e.g., ``There was a significant increase in mortality ...''; see \cite{lehman2019inferring} for more details.
We enclose these spans in special tags.
e.g., {\tt <population>}Participants were diabetics ... {\tt </population>}.
As additional supervision we run the same extraction models over the targets and also demarcate these using the same set of tags. 


An issue with standard sequence-to-sequence models for this task is that they provide no natural means to assess the provenance of tokens in outputs, which makes it difficult to verify the trustworthiness of generated summaries. 
Next we discuss an alternative architecture which is intended to provide greater transparency and controllability. 

\subsection{Proposed Aspect Structured Architecture to Increase Transparency}\label{sec:structured_model}

We adopt a multi-headed architecture similar to \cite{goyal2021hydrasum}, which explicitly generates tokens corresponding to the respective aspects (Figure \ref{fig:struct-sum}).
We assume inputs are segmented into texts corresponding to a set of $K$ fields or aspects. 
Here these are descriptions of trial populations, interventions, and outcomes, and ``punchline'' snippets reporting the main study findings. 
We will denote inputs for each of the $K$ aspects by \{$x^{a_1} , ... , x^{a_K}$\}, where $x^{a_k}$ denotes the text for aspect $k$ extracted from input $x$.
Given that this is a multi-document setting (each input consists of multiple articles), $x^{a_k}$ is formed by \emph{concatenating aspect texts across all documents}
using special tokens to delineate individual articles.

We encode aspect texts separately to obtain aspect-specific embeddings $x_{\mathrm{enc}}^{a_k}$.
We pass these (respectively) to aspect-specific decoders and a shared language model head to obtain vocabulary distributions $\hat{o}_{t}^{a_{k}}$. 
All model parameters 
are shared save for the last two decoder layers which comprise aspect-specific parameters.
Importantly, the representation for a given aspect is \emph{only based on the text associated with this aspect} ($x^{a_k}$).

We model the final output as a \emph{mixture} over the respective aspect distributions:
 $\hat{o}_{t} = \sum_{k=1}^{K} z_{t}^{a_{k}} (\hat{o}_{t}^{a_{k}})$.
Mixture weights $z_{t} = {z_{t}^{a_{1}},\dots,z_{t}^{a_{K}}}$ encode a soft selection over aspects for timestep $t$ and are obtained as a dot product between each penultimate representation of the decoder $y_{t}^{a_k}$ (prior to passing them through a language model head) and a learnable parameter, $W_{z} \in R^{D}$. 
The $K$ logits $\tilde{z}_{t}^{a_{k}}$ are then normalized via a Softmax before multiplying with the aspect-specific vocabulary distributions $\hat{o}_{t}^{a_{k}}$

\vspace{.3em}
\noindent\textbf{Tracing outputs to inputs} This architecture permits one to inspect the mixture weights associated with individual tokens in a generated summary, which suggests which aspect (most) influenced the output. 
Further inspection of the corresponding snippets from studies for this aspect may facilitate verification of outputs, and/or help to resolve errors
and where they may have been introduced.

\vspace{.3em}
\noindent\textbf{Controlled generation}
Neural summarization models often struggle to appropriately \emph{synthesize} conflicting evidence to arrive at the correct overall determination concerning a particular intervention effectiveness. 
But while imperfect, summarization models may be useful nonetheless by providing a means to rapidly draft synopses of the evidence to be edited. 
The multi-headed architecture naturally permits template in-filling, because one can explicitly draw tokens from heads corresponding to aspects of interest. 
In our demo, 
we allow users to toggle between different templates which correspond to different conclusions regarding the overall effectiveness of the intervention in question.
(It would be simple to extend this to allow users to specify their own templates to be in-filled.)

\begin{figure}
    \centering
    \includegraphics[scale=.78]{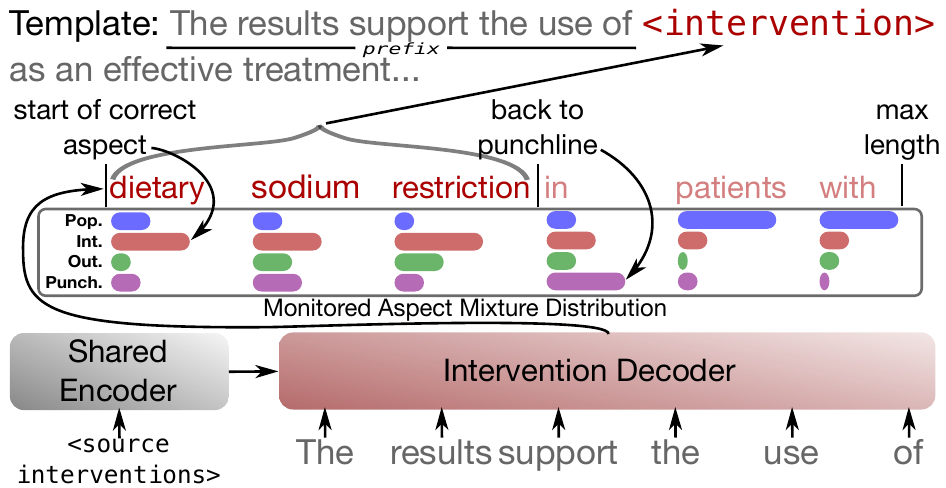}
    \caption{\textbf{Template generation.} To in-fill, we force generation from a specific head and monitor the model's mixture distribution to decide when to stop.}
    \label{fig:template_generation}
\end{figure}

To in-fill templates we use 
template text preceding blanks as context 
and then generate text from the language head corresponding to the designated aspect. 
To determine span length dynamically we monitor the mixture distribution and stop when the it shifts to the another aspect (Figure \ref{fig:template_generation}).

\subsection{User Interface}



\begin{figure*}[tb!]

    \centering
    \includegraphics[scale=.43]{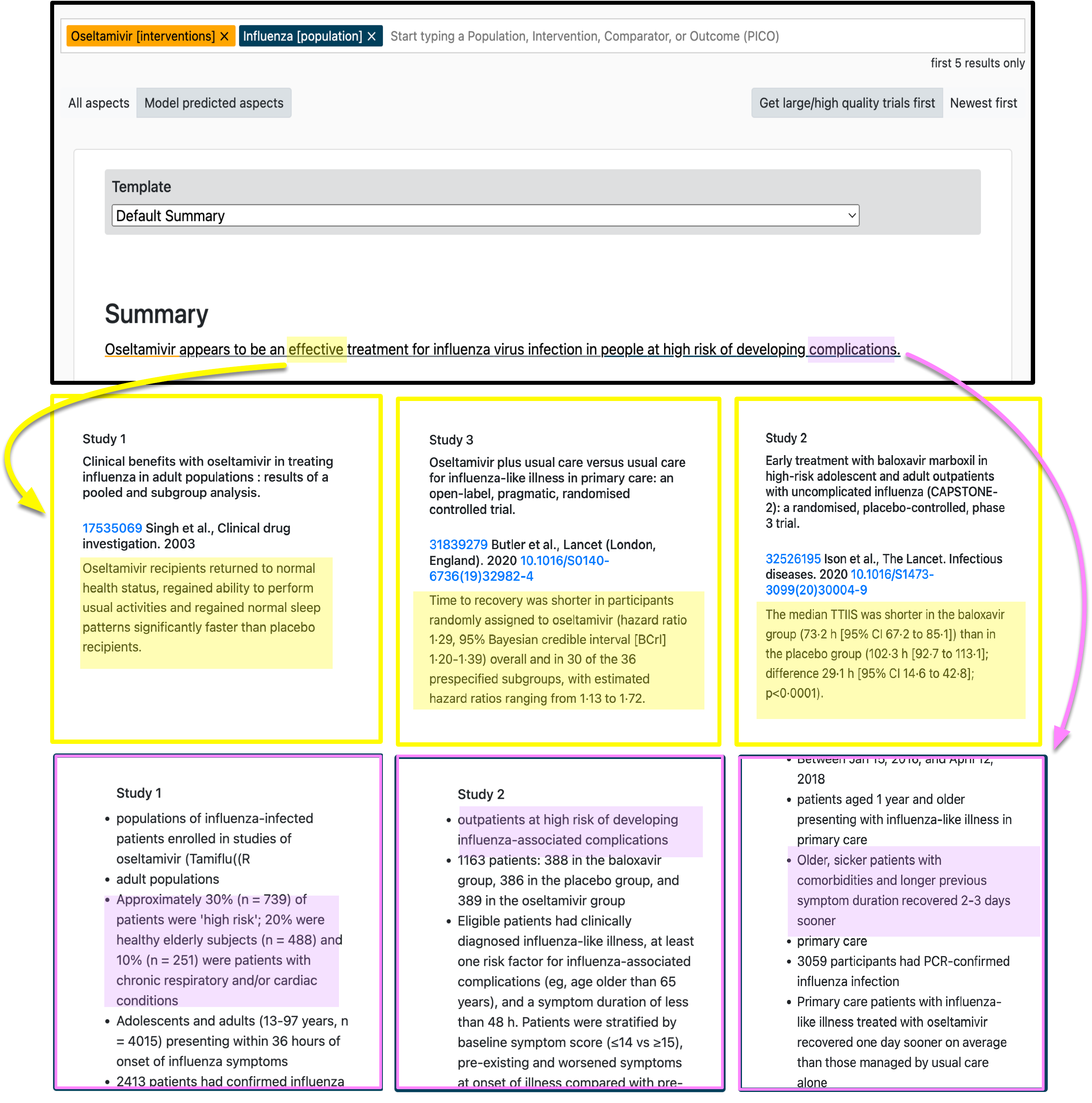}
    \caption{Qualitative example where the structured summarization model (and associated interface) permits token-level verification of the summary generated regarding the use of oseltamivir on influenza-infected patients.
    This approach readily indicates support for the claim that it is ``effective'' (top; yellow) and for the description of the population as individuals at risk of ``complications'' (bottom; purple).}
    
    \label{fig:structured-int-a}
\end{figure*}

Figure \ref{fig:structured-int-a} shows the interface we have built integrating the multi-headed architecture. 
Highlighted aspects in the summary provide a means of interpreting the source of output tokens by indicating the aspects that informed their production. 
One can in turn inspect the snippets associated with these aspects, which may help to identify unsupported content in the generated summary. 
To this end when users click on a token we display the subset of the input that most informed its production. 

We provide additional context by displaying overviews (i.e., ``punchlines'') communicating the main findings of the trials.
Because standard sequence-to-sequence models do not provide a mechanism to associate output tokens with input aspects, we display all aspects (and punchlines) for all trials alongside the summary for this model.

Capitalizing on the aforementioned in-filling abilities of our model, we also provide pre-defined templates for each possible ``direction'' of aggregate findings (significant vs. no effect). We discuss the interface along with examples in Section \ref{sec:case_study}.

\begin{figure*}[tb!]
\centering
    \includegraphics[scale = 0.33]{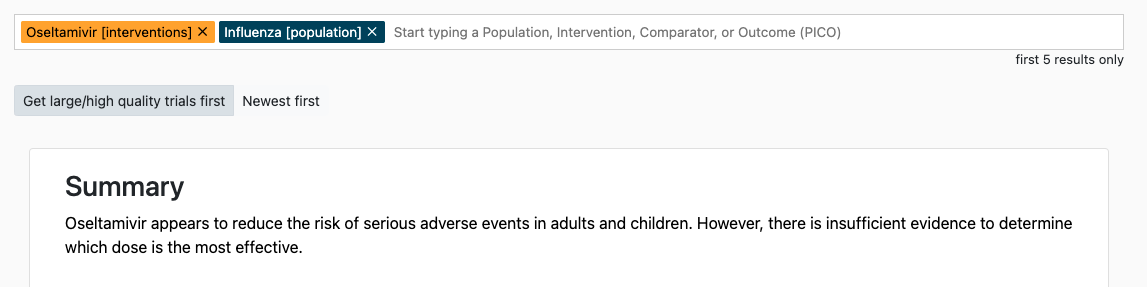}

    \caption{Example output and interface using a standard BART \cite{lewis2019bart} model.}
    \label{fig:structured-vanilla-sum}
\end{figure*}

\section{Dataset and Training Details}
\label{sec:appendix_data_training}
\vspace{-.4em}

We aim to consume collections of titles and abstracts that describe 
RCTs addressing the same clinical question to abstractive summaries 
that synthesize the evidence presented in these. 
We train all models 
on an RCT summarization dataset \cite{wallace2021generating} where we extract clinically salient elements --- i.e., our aspects --- from each of the (unstructured) inputs as a pre-processing step using existing models \citep{marshall2020trialstreamer}.

\vspace{.3em}
\noindent\textbf{Training} We use the Huggingface Transformers library \cite{wolf2020transformers} to implement both models. 
We initalize both models to \textit{bart-base} \cite{lewis2019bart}. 
We fine-tune the models with a batch size of 2 for 3 epochs, using the Adam optimizer \cite{kingma2014adam} with a learning rate of 3e-5. 

\vspace{.3em}
\noindent\textbf{Inference} We use beam search with a beam size of 3. We set the min and max length of generated text to be 10 and 300, respectively.

\section{Case Study: Verification and Controllability}
\label{sec:case_study}
To demonstrate the potential usefulness of the interface (and the architecture which enables it), we walk through two case studies.
We highlight the type of interpretability for verification our proposed approach provides, also demonstrate the ability to perform controllable summarization to show how this might be useful.
The queries used in these case studies along with the investigation were performed by a co-author IJM, a medical doctor with substantial experience in evidence-based medicine. 
We also compare the models and report automatic scores for ROUGE and factuality in the Appendix section \ref{appendix_sec:eval} and find that the two models perform comparably.


\begin{figure*}[tb!]
    \centering
    \includegraphics[scale=.34]{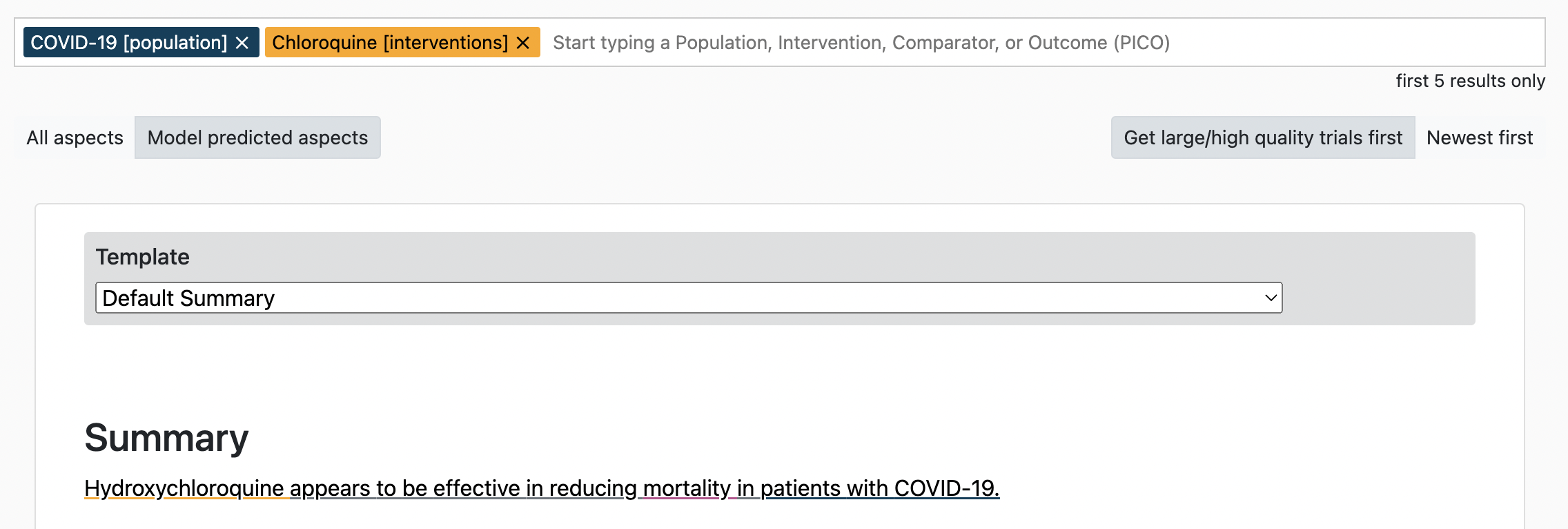}
    
     \centering
    \includegraphics[scale=.34]{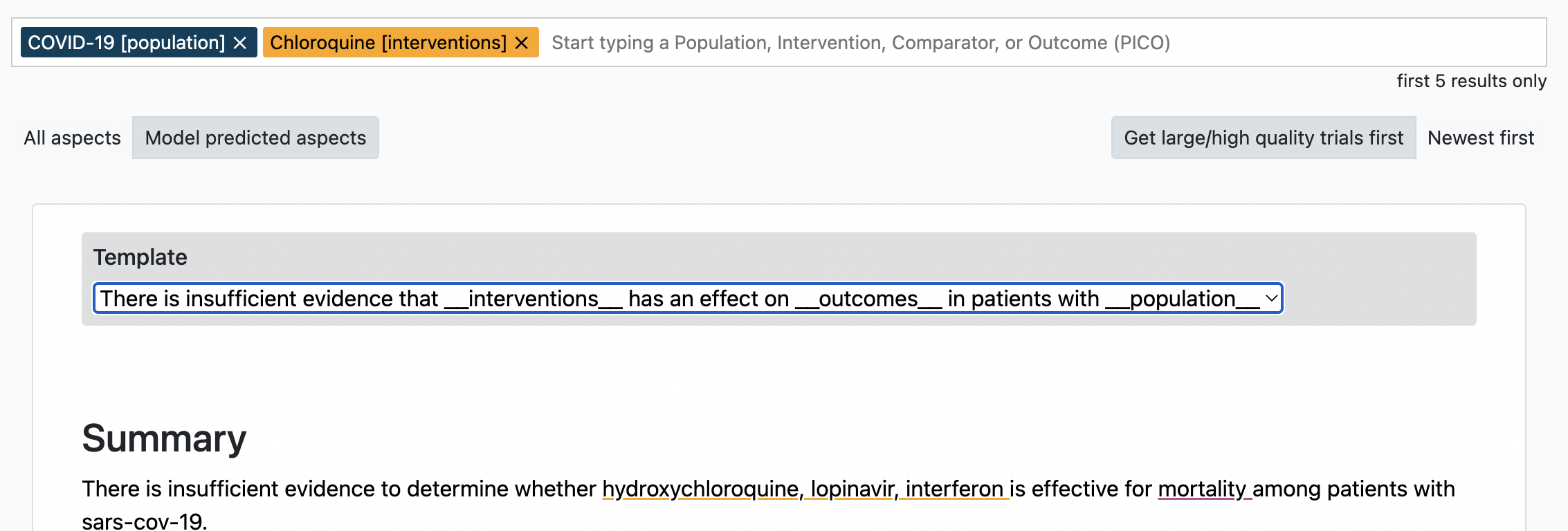}
    
    \caption{Inaccurate summaries generated by the structured model regarding the effect of Chloroquine on patients with COVID-19 (top). Template-controlled summary using the structured model (bottom).}
    \label{fig:incorrect_summaries}
\end{figure*}

\paragraph{Model Interpretability}
As an example to highlight the potential of the proposed architecture and interface to permit verification, we consider a query regarding the effect of Oseltamivir as an intervention for patients infected with influenza. 
The standard architecture produces a summary of the top most relevant RCTs to this query shown in Figure \ref{fig:structured-vanilla-sum}. This comprises two claims: (1) The intervention has been shown to reduce the risk of adverse events among adults and children, and, (2) There is no consensus as to the most effective dosage. 
One can inspect the inputs to attempt to verify these.
Doing so, we find that reported results do tend to indicate a reduced risk of adverse events and that adolescents and adults were included in some of these studies, indicating that the first claim is accurate.
The second claim is harder to verify on inspection; no such uncertainty regarding dosage is explicitly communicated in the inputs.
Verifying these claims using the standard seq2seq architecture is onerous because the abstractive nature of such models makes it difficult to trace parts of the output back to inputs.
Therefore, verification requires reading through entire inputs to verify different aspects. 


The multi-headed architecture allows us to provide an interactive interface intended to permit easier verification.
In particular, associating each output token with a particular aspect provides a natural mechanism for one to inspect snippets of the inputs that might support the generated text. 
Figure \ref{fig:structured-int-a} illustrates this for the aforementioned Oseltamivir and flu example.
Here we show how the ``effective'' token in the output can be clicked on to reveal the aspect that influenced its production (Figure \ref{fig:struct-sum}), in this case tracing back to the extracted ``punchlines'' conveying main study findings.
This readily reveals that the claim is supported.
Similarly, we can verify the bit about the population being individuals at risk of complications by tracing back to the population snippets upon which this output was conditioned. 

\paragraph{Controllability} As mentioned above, another potential benefit of the proposed architecture is the ability to ``in-fill'' templates to imbue neural generative models with controllability. 
In particular, given that the overall (aggregate) treatment efficacy is of primary importance in this context, we pre-define templates which convey an effect direction.
The idea is that if upon verification one finds that the model came to the wrong aggregate effect direction, they can use a pre-defined template corresponding to the correct direction to generate a more accurate summary on-demand.


We show an example of a summary generated by the structured model in the top part of Figure \ref{fig:incorrect_summaries}. 
By using the interpretability features for verification discussed above, we find that the model inaccurately communicates that the intervention Chloroquine is effective for treating COVID-19. 
However, with the interactive interface we are able to immediately generate a new summary featuring the corrected synthesis result (direction), as depicted in the bottom of Figure \ref{fig:incorrect_summaries}, without need for manual drafting.

We provide additional case studies in Appendix Section \ref{appendix_sec:case_studies}.
\section{Conclusions}
We have described TrialsSummarizer, a prototype system for automatically summarizing RCTs relevant to a given query.
Neural summarization models produce summaries that are readable and (mostly) relevant, but their tendency to introduce unsupported or incorrect information into outputs means they are not yet ready for use in this domain.

We implement a multi-headed architecture intended to provide greater transparency.
We provided qualitative examples intended to highlight its potential to permit faster verification and controllable generation. 
Future work is needed to test the utility of this functionality in a user trial, and to inform new architectures that would further increase the accuracy and transparency of models for summarizing biomedical evidence.

\section*{Limitations and Ethical Issues}

\paragraph{Limitations} This work has several limitations.
First, as stated above, while the prospect of automatic summarization of biomedical evidence is tantalizing, existing models are not yet fit for the task due to their tendency to introduce factual errors.
Our working prototype serves in part to highlight this and motivate work toward resolving issues of reliability and trusworthiness. 

In this demo paper we have also attempted to make some progress in mitigating such issues by way of the proposed structured summarization model and accompanying interface and provided qualitative examples highlighting its potential, but really a formal user study should be conducted to assess the utility of this. 
This is complicated by the difficulty of the task: To evaluate the factuality of automatic summaries requires deep domain expertise and considerable time to read through constituent inputs and determine the veracity of a generated summary.


Another limitation of this work is that we have made some ad-hoc design decisions in our current prototype system. 
For example, at present we (arbitrarily) pass only the top-5 (based on trial sample size and estimated reliability) articles retrieved for a given query through the summarization system.
Future work might address this by considering better motivated methods to select which and how many studies ought to be included.



\paragraph{Ethics}

Accurate summaries of the biomedical evidence have the potential to ultimately improve patient care by supporting the practice of evidence-based medicine.
However, at present such models bring inherent risks.
In particular, one may be tempted to blindly trust model outputs; given the limitations of current summarization technologies, this would be ill-advised. 

Our prototype demonstration system is designed in part to highlight existing challenges that must be solved in this space before any model might actually be adopted (and beyond this, we emphasize that need for verification of outputs, which has been the focus of the present effort). In the interface we indicate with a hard-to-miss warning message that this system should only be used for research purposes and these summaries are unreliable and \emph{not to be trusted}.



\section*{Acknowledgements}

This work was supported in part by the National Institutes of Health (NIH) under award R01LM012086, and by the National Science Foundation (NSF) awards 1901117 and 2211954.
The content is solely the responsibility of
the authors and does not necessarily represent the
official views of the NIH or the NSF.

\bibliography{anthology,custom}
\bibliographystyle{acl_natbib}

\appendix

\section*{Appendix}

\section{Automatic Evaluation}\label{appendix_sec:eval}
We report ROUGE scores with respect to the target
(manually composed) Cochrane summaries, for both the development and test sets. 
We report scores for both the vanilla standard BART model along with our proposed multi-headed model intended to aid verifiability and controllability. 
The models perform about comparably with respect to this metric as can be seen in Table \ref{tab:rouge}. 

However ROUGE measures are based on (exact) n-gram overlap, and cannot measure the factuality of generated texts. 
Measuring factuality is in general an open problem, and evaluating the factual accuracy of biomedical reviews in particular is further complicated by the complexity of the domain and texts.
Prior work has, however, proposed automated measures for this specific task \cite{wallace2021generating,deyoung2021ms2}.
These metrics are based on models which infer the reported \emph{directionality} of the findings, e.g., whether or not a summary indicates that the treatment being described was effective.
More specifically, we make binary predictions regarding whether generated and reference summaries report significant results (or not) and then calculate the F1 score of the former with respect to the latter.

\begin{table}[h!]
    \centering
    \centering
    \begin{tabular}{c|c|c}
    \hline
         Model & ROUGE-L (dev) &  ROUGE-L(test)\\
    \hline
    BART & 20.4 & 19.7 \\
    Multi-head & 19.9 & 19.3\\
    \hline
        
    \end{tabular}
    \caption{ROUGE scores achieved by the standard BART model and our proposed multi-headed architecture on the dev and test sets.}
    \label{tab:rouge}
\end{table}

\begin{table}[h!]
  \centering
    \begin{tabular}{c|c|c}
    \hline
         Model & Direc (dev) &  Direc(test)\\
    \hline
    BART & 49.6 & 51.8 \\
    Multi-head & 49.3 & 52.7 \\
    \hline
    \end{tabular}
    \caption{Directionality scores on the vanilla BART model and our proposed multi-headed architecture on the dev and test sets.}
    \label{tab:directionality}  
\end{table}

\section{Additional Case Studies}\label{appendix_sec:case_studies}

\begin{figure*}
\centering
    \frame{\includegraphics[scale=0.40]{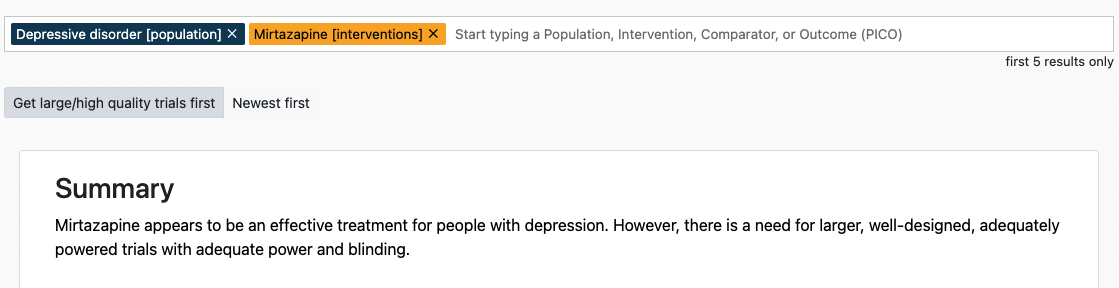}}
    
   \vspace{0.5cm}
   
    \frame{\includegraphics[scale=0.40]{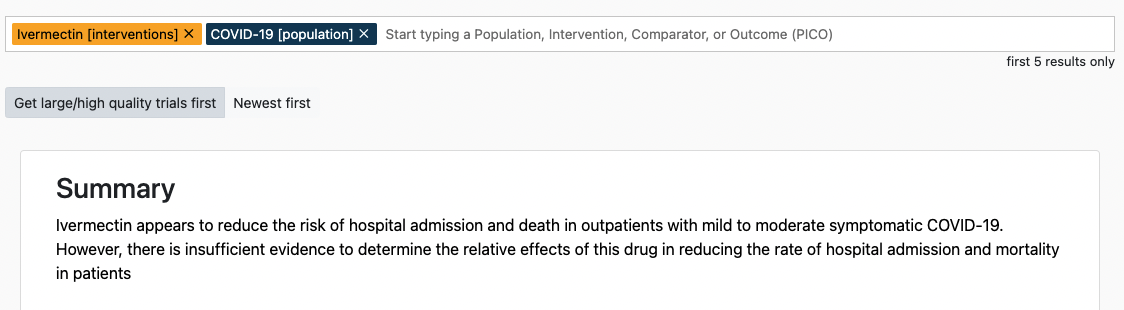}}
    \vspace{0.5cm}
    
    \frame{\includegraphics[scale=0.40]{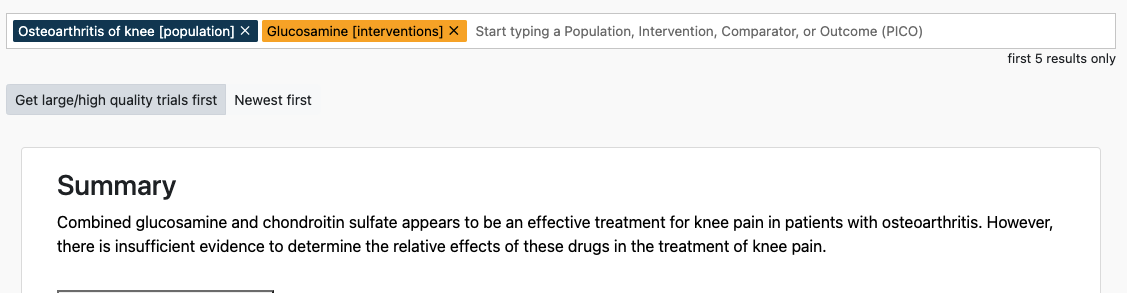}}
    
    \caption{a) BART generated summary when queried about the use of \emph{Mirtazapine} to treat  \emph{depression}
    b) BART generated summary when queried about the use of \emph{Ivermectin} to treat  \emph{COVID-19}) }
    \label{fig:fig-vanilla-cs}

\end{figure*}

\begin{figure*}[tb!]
    \centering
    \frame{\includegraphics[scale=0.42]{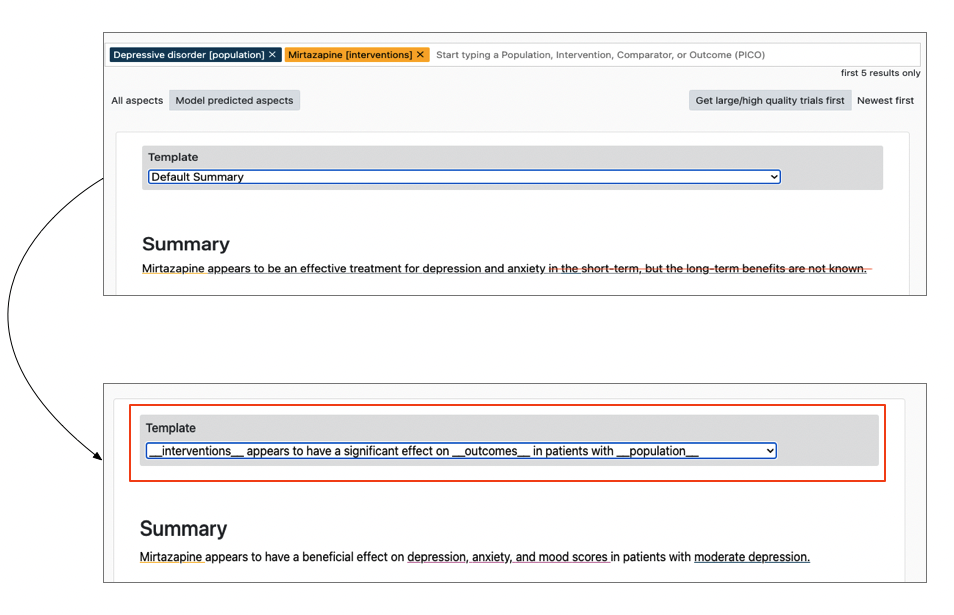}}
    \caption{The summary on top shows the default summary generated by the multi-headed model when queried for the effect of \emph{Mirtazapine} on \emph{depression}. The bottom summary shows the controlled summary using a pre-defined template.}
    \label{fig:multilabel_cs-1}
\end{figure*}

In this section we highlight a few more use cases that demonstrate the need for interpretability and controllability. 

\paragraph{Interpretability} We first highlight a set of examples where verifying model generated summaries is difficult without an interface explicitly designed to provide interpretability capabilities. 
In Figure \ref{fig:fig-vanilla-cs} (a) we show an example where the model generates a summary that accurately synthesized a summary on the effect of using Mirtazapine for patients with depression. However, the summary also includes a statement that states the need for adequate, well-designed trials. Because this statement is generic and does not point to discussing any of the PICO elements, it is unclear what element was responsible for the generation of the statement. A user would therefore need to review all (raw) input texts.


In the case of Figure \ref{fig:fig-vanilla-cs} (b), the model generated summaries has two contradicting sentences. 
The first sentence indicates a reduction in hospital admission and death among COVID-19 patients when Ivermectin was used and the second sentence claims there is insufficient evidence for the same. However without interpretability capabilities it is not possible to debug and verify if the same set of elements were responsible for contradicting statements or not.

The example in Figure \ref{fig:fig-vanilla-cs} (c) shows a case where the model first accurately synthesizes the findings in the studies of the effect of glucosamine in combination of chondroitin sulfate on knee pain. 
However, the following statement talks about the relative effects of the two. 
Again, in this case it is is not intuitive which element led to the generation of the statement and verification requires careful reviewing of all the text and their implication in all elements.

\paragraph{Controllability} We next highlight examples where one can effectively control the generation of summaries that would otherwise be incorrect using the template in-filling capabilities afforded by our model.
While the interpretability features may permit efficient verification, models still struggle to consistently generate factual accurate summaries. 
We showcase instances where one can arrive at more accurate summaries quickly via the controllability (template in-filling) made possible by our model.

In the example shown in Figure \ref{fig:multilabel_cs-1} the default summary  synthesizes the effect accurately. 
However, the model summary discusses the effect on short-term and long-term benefits generated from the punchlines of the studies. 
Reading through extracted `punchlines', we find that the studies indicate issues upon withdrawal but do not necessarily provide information on long-term use of the medication. 
In-filling templates constrains the output, and can be used to produce more accurate summaries while still taking some advantage of the flexibility afforded by generation. 
For instance in this case we can see that the edited summary induced using the template is more accurate.

\begin{figure*}[h!]
    \centering
    \frame{\includegraphics[scale=0.42]{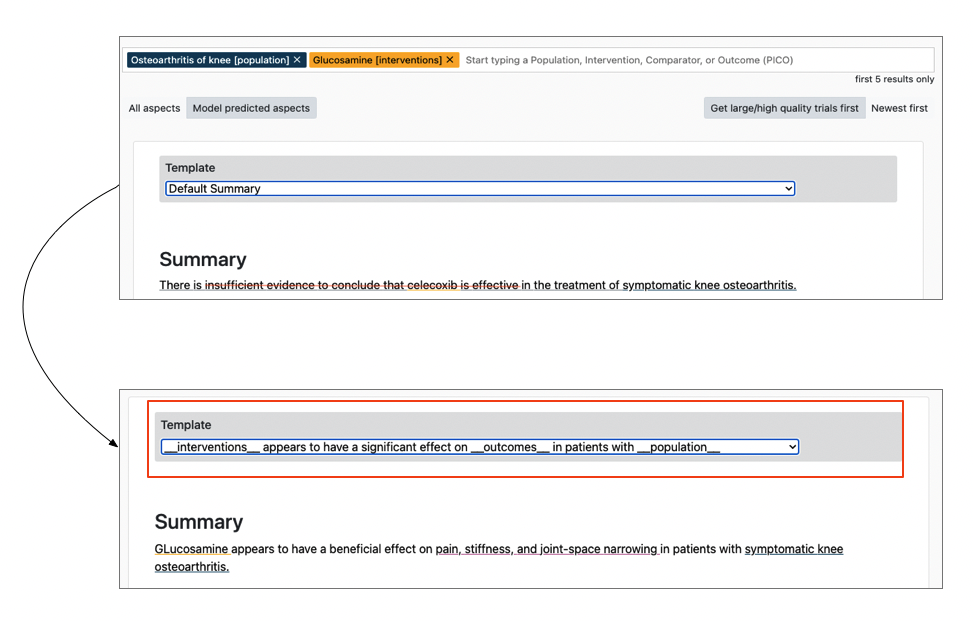}}
    \caption{The summary on top shows the default summary generated by the multi-headed model when queried for the effect of \emph{Glucosamine} on \emph{Osteoarthritis of knee}. The bottom summary shows the edited summary using a pre-defined template}
    \label{fig:multilabel_cs-2}
\end{figure*}

Similarly, in Figure \ref{fig:multilabel_cs-2} when the multi-headed model is queried for the effect of Glucosamine on Osteoarthritis of knee, we observe that the model on its own produces a summary conveying an incorrect aggregate effect of studies.
We can verify this by inspecting the elements responsible for the generation, as discussed above. 
We then arrive at a more accurate summary using the template shown. 

\begin{figure*}[h!]
    \centering
    \frame{\includegraphics[scale=0.42]{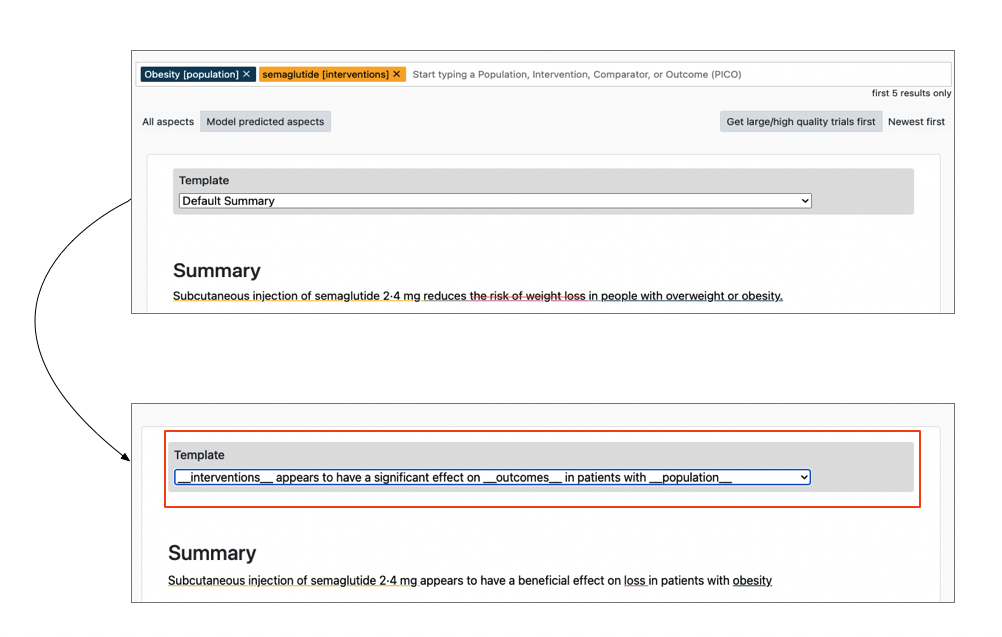}}
    \caption{The summary on top shows the default summary generated by the multi-headed model when queried for the effect of \emph{Semaglutide} on \emph{obese} patients. The bottom summary shows the edited summary using a pre-defined template}
    \label{fig:multilabel_cs-3}
\end{figure*}

The example in Figure \ref{fig:multilabel_cs-3} is an interesting mistake made by the model. 
Because the outcomes can be presented with the same information but in a positive or negative direction (e.g., weight loss vs weight gain), the model has to accurately infer the effect of all studies. 
In this case, the model generates a summary with the right effect but views weight loss as an undesirable effect. 
Here again we select a template and allow the model quickly in-fill, yielding a more accurate summary.

\end{document}